\documentclass[10pt,twocolumn,letterpaper]{article}

\usepackage{wacv}
\usepackage{times}
\usepackage{epsfig}
\usepackage{graphicx}
\usepackage{amsmath}
\usepackage{amssymb}

\usepackage{algorithm}
\usepackage{algorithmic}
\usepackage{diagbox}

\def\wacvPaperID{****} 

\wacvfinalcopy 

\ifwacvfinal
\def\assignedStartPage{1} 
\fi

\ifwacvfinal
\usepackage[breaklinks=true,colorlinks,bookmarks=false]{hyperref}
\else
\usepackage[pagebackref=true,breaklinks=true,colorlinks,bookmarks=false]{hyperref}
\fi

\ifwacvfinal
\else
\fi

\begin{document}

\title{Adversarially Optimized Mixup for Robust Classification}

\author{Jason Bunk, Srinjoy Chattopadhyay, B.S.Manjunath, Shivkumar Chandrasekaran\\
Mayachitra Inc.\\
5266 Hollister Ave, STE 229, Santa Barbara, CA 93111, USA\\
{\tt\small \{bunk,srinjoy,manj,shiv\}@mayachitra.com}
}

\maketitle

\begin{abstract}
\vspace{-3mm}
Mixup \cite{zhang2018mixup} is a procedure for data augmentation that trains networks to make smoothly interpolated predictions between datapoints. Adversarial training \cite{goodfellow2015explaining},\cite{madry2018towards} is a strong form of data augmentation that optimizes for ``worst-case'' predictions in a compact space around each datapoint, resulting in neural networks that make much more robust predictions. In this paper, we bring these ideas together by adversarially probing the space between datapoints, using projected gradient descent (PGD). 
The fundamental approach in this work is to leverage backpropagation through the mixup interpolation during training to optimize for places where the network makes unsmooth and incongruous predictions. Additionally, we also explore several modifications and nuances, like optimization of the mixup ratio and geometrical label assignment, and discuss their impact on enhancing network robustness. Through these ideas, we have been able to train networks that robustly generalize better; experiments on CIFAR-10 and CIFAR-100 demonstrate consistent improvements in accuracy against strong adversaries, including the recent strong ensemble attack AutoAttack\cite{croce2020reliable}. Our source code would be released for reproducibility.
\end{abstract}


\vspace{-4mm}
\section{Introduction}
\label{sec:intro}

The vulnerability of neural networks to adversarial attack has been plaguing machine learning researchers ever since the discovery by Szegedy et al. \cite{szegedy2014intriguing}. In the years since, many research efforts have been geared towards making neural networks robust to adversarial perturbations, but many defense strategies have failed to stand the test of time. One of the strongest baselines for adversarial robustness that has repeatedly stood up to rigorous scrutiny is adversarial training \cite{goodfellow2015explaining}, in particular adversarial training based on the Projected Gradient Descent (PGD) attack strategy \cite{madry2018towards}. PGD adversarial training can be mathematically represented as a constrained inner min-max optimization, whose solution gives us a minimal perturbation that maximizes classification loss. This min-max optimization is a characteristic of many approaches to adversarial robust training, including in other related formulations of the loss, like TRADES \cite{zhang2019theoretically}.

Such networks have repeatedly withstood numerous evaluations, as shown in works like \cite{athalye2018obfuscated}. One concern is that the robust accuracy of such networks on test sets leaves much to be desired. For instance, state of the art neural networks typically have less than 50\% accuracy-under-attack for the CIFAR-10 dataset. This low robust accuracy occurs despite the fact that the network is able to memorize the \textit{robust} ($\ell_\infty$-bounded boxed) training distribution with nearly 100\% accuracy against its own attack model, as a result of PGD adversarial training. This points to a familiar problem for machine learning practitioners -- over-fitting to the training set. As recent work has pointed out \cite{wong2020overfitting}, robust adversarial training is just as susceptible to ``overfitting'' as standard neural network training, because the robust training distribution is an incomplete subsampling or imperfect match to the robust test distribution due to the finite size of the dataset. 

Recent developments have also shown that adversarial vulnerability is a problem of networks severely overfitting to features that are imperceptibly small yet useful for classification \cite{ilyas2019adversarial}. With this view, adversarial training is an advanced form of worst-case data augmentation. This view has been shown in \cite{xie2020adversarial} to be able to usefully improve accuracy under various corruptions including ``natural adversarial examples'' \cite{hendrycks2019natural} with certain domain adaptation tools. Hence, improvements in adversarial training are useful not just for security purposes, but for a wide audience who desires robust classification to be more understandably as well as consistently generalizeable.

Our approach focuses on the poor and overfitting accuracy-under-attack of robust training. We frame together the effective mixup augmentation \cite{zhang2018mixup} introduced for typical (non-adversarial) training, and robust PGD adversarial optimization. We use adversarial optimization to pinpoint the locations in the interpolation space between datapoints, where the classification decisions of the neural networks are the least smooth, i.e. that adversarially maximize the KL divergence between the network's predictions and the smoothed label interpolation between datapoints.

Robust adversarial learning is susceptible to overfitting, and we show that data augmentation insights from standard training transfer well to adversarial optimization. Our contributions include:
\begin{itemize}
\item We show through intuitive geometry and empirical results that previous works integrating mixup and adversarial optimization were limited in their ability to probe the vicinal distribution to find worst-case points. It is important for the adversarial optimization to be able to fully find the worst-case points to learn from.
\item With ablation experiments we break down the optimization components that led our results to surpass the baselines.
\item Our approach demonstrates significant improvements in robust accuracy-under-attack against strong, state-of-the-art adversaries. We evaluate against an ensemble of state-of-the-art adversaries including a strong gradient-free black-box attack, demonstrating that our approach provides real improvements that do not introduce or rely on any gradient obfuscation.
\end{itemize}

The rest of this paper is organized as follows. In Section \ref{sec:relworks}, we summarize related work exploring mixup for data augmentation or adversarial robustness. We present the background of this work and the details of our approach in Section \ref{sec:approach}. Section \ref{sec:exp} presents the experimental results of our work, which are then comprehensively discussed in Section \ref{sec:discussion}. Finally, we conclude this paper and discuss future avenues of research in Section \ref{sec:conc}.



\section{Related Work}
\label{sec:relworks}

The concept of Mixup \cite{zhang2018mixup} was initially proposed as an introduction to Vicinal Risk Minimization to neural network training, that reduces overfitting by encouraging the network to make smooth, linearly interpolated predictions between datapoints. Manifold Mixup \cite{verma2019manifold} is a recent update that also performs such interpolations between intermediate features in the network, encouraging features throughout the model to smoothly interpolate between datapoints. However, neither approach is robust to multi-step adversarial attacks. Another related approach, Adversarial Vertex Mixup \cite{lee2020adversarial} considers first finding the adversarial perturbation point $X_{av}$ for each datapoint $X$, then using Mixup to mix $X$ and $X_{av}$ for the purpose of learning a label-smoothed distribution around $X$. However, this method does not make use of the valuable space \textit{between} different datapoints that can be learned with the original Mixup \cite{zhang2018mixup}, which is one of the key ideas behind the work presented in this paper. 

Wong et al. \cite{wong2020overfitting} evaluated adversarial training with mixup and found reduced performance compared to a well-tuned PGD-only baseline. We refer to their implementation as the Baseline Algorithm \ref{algo:baseline1} (see Fig. \ref{figbaseline1}) and indeed found that it is not particularly effective. Mixup Inference \cite{pang2020mixup} uses Mixup in the inference phase by using minibatches at test time to improve adversarial robustness. However, concerns with this approach have been raised by \cite{tramer2020adaptive}. In our work, at inference time we do nothing but a standard deterministic prediction, which avoids such concerns. Guo et al. \cite{guo2019mixup} learn a prior distribution for the mixup ratio using the reparameterization trick, but their approach is not oriented for adversarial robustness. As a result, similarly to baseline Mixup and Manifold Mixup, this strategy will be vulnerable to worst-case perturbations.

Directional Adversarial Training \cite{archambault2019mixup} has been proposed for adversarially optimizing the mixup ratio between randomly sampled pairs. However, their approach will miss the space \textit{around} points (in small $\ell_\infty$-bound boxes) that adversarial robustness focuses on (in the same way that ordinary mixup is not robust to adversarial attack), and they do not try to evaluate for this case.


VarMixup \cite{mangla2020varmixup} uses a formulation of adversarial mixup to improve their generative VAE model, but their strongest results on adversarial classification are dependent upon Mixup Inference \cite{pang2020mixup}, the value of which has been disputed by Tramer et al \cite{tramer2020adaptive} who were able to attack through Mixup Inference to be no better than baseline PGD adversarial training. To avoid such concerns, we avoid using any inference-time ``tricks'' and evaluate our networks the same as the baseline PGD adversarial training \cite{madry2018towards}.


\vspace{-1mm}
\section{Approach}
\label{sec:approach}
In this section, we establish the required background and discuss the necessary details of our approach. We frame the threat model we are aiming to defend. Then, we walk through two baseline training schemes that use mixup and adversarial training, before introducing our approach, in order to show by comparison the novelties of our approach. Additionally, we discuss a few nuances and optimization tools in our geometrical framework. 
\vspace{-1mm}
\subsection{Threat Model}
We start this section by discussing the adversarial threat model considered in this work. We work with the threat model of $\ell_\infty$ norm bounded attacks, and use white-box evaluation. Our models are trained using adversarial optimization with an inner loop to find a worst-case data augmentation. Thereafter, at test time we transparently use the trained network weights for deterministic predictions. This follows the threat model of \cite{madry2018towards}. We use PGD adversarial training from \cite{madry2018towards} detailed as Algorithm \ref{algo:PGD} as our inner optimizer. After producing the adversarial images $x' = (x+\delta)$, the network is trained to make predictions on such images with a standard cross-entropy loss.

\begin{algorithm}
\caption{PGD Adversarial Optimization}
\label{algo:PGD}
\begin{algorithmic}
\REQUIRE { $f_{\theta}(\cdot)$: Neural network with parameters $\theta$ }
\REQUIRE { $\mathcal{L}(\cdot, \cdot)$: Loss (KL divergence)}
\REQUIRE { $D$: Training data with images $x$ and labels $y$ }
\REQUIRE { $\epsilon$: $\ \ell_\infty$ norm bound to constrain adversarial attack }
\REQUIRE { $\eta$: step scale for PGD update }
\FOR { $\{(x,y)\} \sim D $ }
\item $\delta \leftarrow U(-\epsilon,\epsilon) \quad\triangleright$ initialize adversarial perturbation
\FOR { PGD step }
\item $ \mathcal{L}_P = d(f(x+\delta),y) $
\item $ g_\delta \leftarrow \nabla_\delta \mathcal{L}_P \quad\triangleright $ backpropagate gradients
\item $ \delta \leftarrow $ clamp $( \delta + sign(g_\delta) \cdot \eta, -\epsilon, \epsilon) $
\ENDFOR

\item $ \mathcal{L}_a = d(f(x+\delta),y) $
\item $ g_\theta \leftarrow \nabla_\theta \mathcal{L}_a \quad\triangleright $ backpropagate gradients
\item $ \theta \leftarrow $ Step$(\theta,g_\theta) \quad\triangleright$ parameter update

\ENDFOR
\end{algorithmic}
\end{algorithm}

\vspace{-1mm}
\subsection{Baselines}
\begin{figure}[!t]
\centering
\includegraphics[width=2.5in]{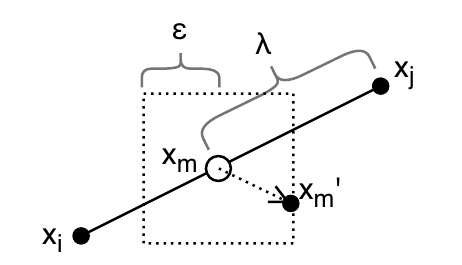}
\caption{Baseline: Mixup, then Attack. See Algorithm \ref{algo:baseline1}. Evaluated by \cite{wong2020overfitting}. The dotted lines indicate the PGD adversarial optimization procedure: the adversarial perturbation is constrained by $\ell_\infty$ norm to within the $\varepsilon$-box around $x_m$.}
\label{figbaseline1}
\end{figure}

We first describe two baseline approaches that combine adversarial training and Mixup for producing robust classifiers. The first approach, starts with choosing a mixing ratio $\lambda$ between two datapoints $(x_i,y_i)$ and $(x_j,y_j)$ and interpolates between them using $\lambda$ to form a virtual datapoint $(x_m,y_m)$. Following this, an adversarial perturbation $x_m'$ is optimized while being constrained around $x_m$, which is then used to train the network with the loss $L(f(x_m'),y_m)$. This is the formulation evaluated in \cite{wong2020overfitting}. An intuitive explanation of this approach is presented in Fig. \ref{figbaseline1}.

\begin{algorithm}
\caption{Baseline: Mixup, Then Attack; as in \cite{wong2020overfitting}}
\label{algo:baseline1}
\begin{algorithmic}
\REQUIRE { $f_{\theta}(\cdot)$: Neural network with parameters $\theta$ }
\REQUIRE { $\mathcal{L}(\cdot, \cdot)$: Loss (KL divergence)}
\REQUIRE { $D$: Training data with images $x$ and labels $y$ }
\REQUIRE { $\varepsilon$: $\ \ell_\infty$ norm bound to constrain adversarial attack }
\FOR { $\{(x_i,y_i),(x_j,y_j)\} \sim D $ }
\item $\lambda \sim B(\alpha, \alpha)$
\item $x_m = \lambda x_i + (1-\lambda) x_j \quad\triangleright $ mixup
\item $y_m = \lambda y_i + (1-\lambda) y_j \quad\triangleright $ mixup
\item $x_m' \leftarrow attack(x_m,y_m,\varepsilon) \quad\triangleright $ adversarial attack (PGD)
\item $ \mathcal{L}_a = d(f(x_m'),y_m) $
\item $ g_\theta \leftarrow \nabla_\theta \mathcal{L}_a \quad\triangleright $ backpropagate gradients
\item $ \theta \leftarrow $ Step$(\theta,g_\theta) \quad\triangleright$ parameter update
\ENDFOR
\end{algorithmic}
\end{algorithm}

\begin{figure}[!t]
\centering
\includegraphics[width=3.3in]{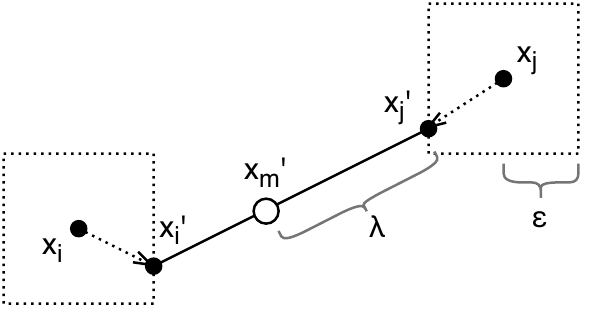}
\caption{Baseline: Adversarially attack, then Mixup. This is the adversarial optimization approach of ``Interpolated Adversarial Training'' \cite{lamb2019interpolated}. The dotted lines indicate the PGD adversarial optimization procedure: the adversarial perturbations are constrained by $\ell_\infty$ norm to within the $\varepsilon$-boxes around $x_i$ and $x_j$.}
\vspace{-2mm}
\label{figbaseline2}
\end{figure}

For the second approach detailed in Fig. \ref{figbaseline2}, the datapoints $(x_i,y_i)$ and  $(x_j,y_j)$ are adversarially perturbed to $(x_i',y_i)$ and $(x_j',y_j)$, respectively. This adversarial perturbation is done independently via the PGD optimization. After this step, we interpolate between $x_i'$ and $x_j'$ to form $x_m'$. The network is again trained with the loss $L(f(x_m'),y_m)$. 

\begin{algorithm}
\caption{Baseline: Attack, then Mixup: Interpolated Adversarial Training (ignoring the unperturbed loss $\mathcal{L}_c$) \cite{lamb2019interpolated}}
\label{algo:interpadvtrain}
\begin{algorithmic}
\REQUIRE { $f_{\theta}(\cdot)$: Neural network with parameters $\theta$ }
\REQUIRE { $\mathcal{L}(\cdot, \cdot)$: Loss (KL divergence)}
\REQUIRE { $D$: Training data with images $x$ and labels $y$ }
\REQUIRE { $\varepsilon$: $\ \ell_\infty$ norm bound to constrain adversarial attack }
\FOR { $\{(x_i,y_i),(x_j,y_j)\} \sim D $ }
\item $x_i' \leftarrow attack(x_i,y_i,\varepsilon) \quad\triangleright $ adversarial attack (PGD)
\item $\lambda \sim B(\alpha, \alpha)$
\item $x_m' = \lambda x_i' + (1-\lambda) x_j' \quad\triangleright $ mixup
\item $y_m = \lambda y_i + (1-\lambda) y_j \quad\triangleright $ mixup
\item $ \mathcal{L}_a = d(f(x_m'),y_m) $
\item $ g_\theta \leftarrow \nabla_\theta \mathcal{L}_a \quad\triangleright $ backpropagate gradients
\item $ \theta \leftarrow $ Step$(\theta,g_\theta) \quad\triangleright$ parameter update
\ENDFOR
\end{algorithmic}
\end{algorithm}

In the paper \cite{lamb2019interpolated}, it is advocated to train the neural network using the sum of the adversarial loss $\mathcal{L}_a$ from Baseline Algorithm \ref{algo:interpadvtrain} and the (mixup) loss on the pristine data $\mathcal{L}_c(f(x_m),y_m)$ where $(x_m,y_m)$ are computed as in Baseline Algorithm \ref{algo:baseline1}. Here, we focus on the adversarial optimization procedure  targeting a worst-case loss comparable to $\mathcal{L}_a$, so as to not preclude the use of pristine (unperturbed) losses.
\vspace{-2mm}
\subsection{Integrated Adversarial Optimization \& Mixup}

\begin{figure}
\centering
\includegraphics[width=3.2in]{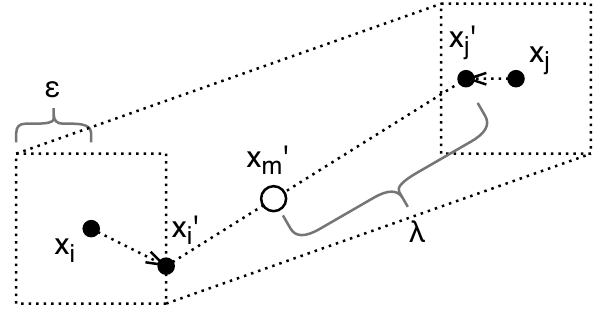}
\caption{Our integrated adversarial mixup optimization. The dotted lines indicate our PGD optimization procedure: the adversarial perturbations are constrained by $\ell_\infty$ norm to within the $\varepsilon$-boxes around $x_i$ and $x_j$. Though we constrain the optimization to the boxes around $x_i$ and $x_j$, the PGD optimizer does not care about $f(x_i')$ or $f(x_j')$; its goal is to use $x_i'$ and $x_j'$ to tug on the wire that connects them, to find the point $x_m'$ in the whole space between that causes $f(x_m')$ to make an incongruous prediction (away from $y_m$). In Algorithm \ref{algo:ouroptimizeratio}, $\lambda$ is added to the adversarial optimization to better explore this interpolation space to find the worst $x_m'$. Not shown is how to produce label $y_m$, for which we will later propose distinguishing $\lambda$ for interpolations $\lambda_x$ and $\lambda_y$ for input $x_m'$ and label $y_m$, respectively.}
\vspace{-2mm}
\label{figourmixup}
\end{figure}

Our approach is shown in Fig. \ref{figourmixup} and the initial implementation is detailed in Algorithm \ref{algo:ours}. We integrate mixup into the adversarial optimization. This will allow us to backpropagate and learn the mixing interpolation ratio $\lambda$, which we find to be beneficial. We will point out a geometrical quirk that compels us to develop a geometrical labeling fix.

Fig. \ref{figourmixup} shows geometrically that the volume of data space in which adversarial perturbations is optimized is greater than previous baselines. We argue that this results in a stronger adversarial learning, because the adversary is able to probe worst-case regions in the data space with far greater flexibility. A commonly observed wisdom is that there is a trade-off between adversarial robustness and accuracy \cite{zhang2019theoretically}. With this perspective, since we are building a stronger adversary, we would expect to increase robustness potentially at the cost of (hopefully slightly) decreasing the accuracy on pristine images. In many domains, including security and safety applications, the worst-case behavior is particularly concerning, so our goal is to improve the robustness against worst-case attacks.

\begin{algorithm}
\caption{Adversarially Optimized Mixup}
\label{algo:ours}
\begin{algorithmic}
\REQUIRE { $f_{\theta}(\cdot)$: Neural network with parameters $\theta$ }
\REQUIRE { $\mathcal{L}(\cdot, \cdot)$: Loss (KL divergence)}
\REQUIRE { $D$: Training data with images $x$ and labels $y$ }
\REQUIRE { $\varepsilon$: $\ \ell_\infty$ norm bound to constrain adversarial attack }
\REQUIRE { $\eta$: step scale for PGD update }
\FOR { $\{(x_i,y_i),(x_j,y_j)\} \sim D $ }
\item $\lambda \sim B(\alpha, \alpha)$
\item $\lambda_x = \lambda_y = \lambda $
\item $\delta_i, \delta_j \leftarrow U(-\varepsilon,\varepsilon) \quad\triangleright$ initialize adversarial perturbations
\FOR { PGD step }
\item $x_m' = \lambda_x (x_i+\delta_i) + (1-\lambda_x) (x_j+\delta_j) \quad\triangleright $ mixup
\item $y_m = \lambda_y y_i + (1-\lambda_y) y_j \quad\triangleright $ mixup
\item $ \mathcal{L}_P = d(f(x_m'),y_m) $
\item $ g_\delta \leftarrow \nabla_\delta \mathcal{L}_P \quad\triangleright $ backpropagate gradients
\item $ \delta \leftarrow $ clamp$( \delta + sign(g_\delta) \cdot \eta, \ -\varepsilon,\ \varepsilon) $
\ENDFOR

\item $ \mathcal{L}_a = d(f(x+\delta),y) $
\item $ g_\theta \leftarrow \nabla_\theta \mathcal{L}_a \quad\triangleright $ backpropagate gradients
\item $ \theta \leftarrow $ Step$(\theta,g_\theta) \quad\triangleright$ parameter update

\ENDFOR
\end{algorithmic}
\end{algorithm}
\vspace{-1mm}
\subsection{Independence of $\delta_i$, $\delta_j$}
In most implementations of mixup, including the current paper, datapoint pairs $[(x_i,y_i), (x_j,y_j)]$ come from a minibatch of examples, where $(x_j,y_j)$ is matched as a permutation of the same minibatch. (In some cases, then, by small chance, $(x_i,y_i) = (x_j,y_j)$, which is equivalent to sampling an interpolation ratio $\lambda$ as 0 or 1). This means that when optimizing the perturbations $\delta = (x'-x)$ that are added to the minibatch $x$, the same perturbation would be duplicated on the left and on the right side of the mixing. Yet, in our formulation, the perturbations need to behave as ``puppeteers'' pulling on the interpolation that occurs along the taut string connecting the two adversarial points. For this reason, for the permuted right side, we allocate a copy of the minibatch inputs $\{x\}$ of shape $(M,...)$ for minibatch size $M$, and initialize perturbations $\{\delta\}$ of shape $(2M,...)$. This frees up the puppeteering optimization to find the best interpolation point between all pairs without any duplication interference. In our ablation studies, we refer to the effect of this as Shared $\delta$. It is important to note that our best model does not use the shared $\delta$ but allocates independently initialized $\delta_i$, $\delta_j$.

\vspace{-1mm}
\subsection{Geometric Label Mixing}
In algorithm \ref{algo:ours}, the mixed label $y_m$ is assigned simply as the linear interpolation of the two source labels $y_i$ and $y_j$. However, because the interpolated datapoints $x_i'$ and $x_j'$ are attacked, it is possible for the label interpolation $y_m$ to be ``out-of-sync'' with the inputs. If for example the adversarial perturbation on $x_i$ is in the direction away from $x_j$, then part of the $\lambda$ of \ref{algo:ours} is working to bring the label $y_i$ \textit{back} in the direction of $y_j$, which can be harmful. For example, if by coincidence $ \delta_i = -c (x_j - x_i) $  for some scalar $c > 0$, and $ \delta_j \ll \lvert  x_i - x_j \rvert $, then
\begin{align*}
 x_m' = \lambda (1+c) x_i + (1 - \lambda(1 + c)) x_j
\end{align*}

which would effectively make $\lambda \longrightarrow \lambda (1+c)$ which is counter-productive to the label learning. In a destructive case (perhaps by coincidence) where $c = 1/\lambda - 1$, then the perturbation $\delta_i$ would just bring $x_m' \rightarrow x_i$, implying a ground truth label $y_i$, but the ground truth label $y_m$ would still be interpolated (smoothed) inbetween $y_i$ and $y_j$ since in mixup the label $y_m$ is dependent only on $\lambda$, not on $c$ or perturbations $\delta_i$ or $\delta_j$.

To address this potential counterproductive case, we propose to derive the label mixing ratio $\lambda_y$ using Algorithm \ref{algo:betterlabelassignment}. The formula is the normalized distance of the point $x_m'$ along the vector projection of the line segment from $x_i'$ to $x_j'$. The result is geometrically consistent, and usefully, it is differentiable.


\begin{algorithm}
\caption{Geometrical Mixed Label Assignment}
\label{algo:betterlabelassignment}
\begin{algorithmic}
\REQUIRE { $ (x_i',y_i), (x_j',y_j) $: perturbed data points }
\REQUIRE { $ x_m' $: mixed sample point, as in Algorithm \ref{algo:ours} }

\STATE $v = x_j' - x_i' $
\STATE $p = x_i' - x_m' $
\STATE $\lambda_y = \textrm{clamp}( 1 + (p \cdot v)/(v \cdot v) , 0, 1)$

\STATE Now use this $\lambda_y$ in place of $\lambda$ in Algorithm \ref{algo:ours}.


\end{algorithmic}
\end{algorithm}

\vspace{-2mm}
\subsection{Learning Mixing Ratio $\lambda_x$}
Now that we have a clear geometrical formulation of how to, simultaneously, adversarially perturb the inputs while interpolating in between datapoints with a consistent label interpolation, we can explore adversarially optimizing the interpolation scalar (for the inputs) $\lambda_x$. We explored the approach of Algorithm \ref{algo:ouroptimizeratio}, in which the mix ratio is clipped to a bounded range between 0.5 and 1 (without loss of generality, clipping from 0.5 to 1 simply assigns ``left''/``right'' sides to each mixup pair), using the sigmoid function as a soft clamp mechanism.

\begin{algorithm}
\caption{Optimizing Mixup Ratio (constrained using sigmoid function $\sigma$)}
\label{algo:ouroptimizeratio}
\begin{algorithmic}
\REQUIRE { $ \kappa $: clip bound between 0.5 and 1 for mix ratio }
\REQUIRE { $ \eta_\gamma $: step scale for updates }

\STATE $\lambda_{init} \sim B(\alpha, \alpha)$
\WHILE {min$(\lambda_{init}) < \kappa$}
\item $ \lambda_{init}[\lambda_{init} < \kappa] \sim B(\alpha, \alpha) $
\ENDWHILE

\STATE $ \gamma \leftarrow -log((1-\kappa)/(\lambda_{init}-\kappa) - 1) $


\FOR { PGD step }
\item $\lambda \leftarrow \kappa + (1-\kappa) \sigma(\gamma)$
\item $\ldots \mathcal{L}_P \quad\triangleright$ as in Algorithm \ref{algo:ours}, using $\lambda$
\item $ g_\gamma \leftarrow \nabla_\gamma \mathcal{L}_P \quad\triangleright $ backpropagation
\item $ \gamma \leftarrow \gamma + sign(g_\gamma) \cdot \eta_\gamma $
\ENDFOR

\end{algorithmic}
\end{algorithm}

\section{Experiments}
\label{sec:exp}

We start this section by presenting the details of our experimental setup. These are followed by extensive experimental results presenting the ablation studies and the relative performance of our approaches with those in literature. 
\subsection{Setup}
For the evaluation of the ideas developed in this paper, we perform experiments on the CIFAR-10 and CIFAR-100 datasets. We use the same network and hyperparameters settings for both. We use the PreAct-ResNet-18 \cite{he2016identity} and Wide ResNet-34 \cite{zagoruyko2016wide} architectures, which have been used by many others for evaluating adversarial robustness \cite{zhang2018mixup,wong2020fast,wong2020overfitting,lamb2019interpolated}. Our implementation uses Pytorch \cite{NEURIPS2019_9015}. Our threat model is the $\ell_{\infty}$ bound 8/255.

As noted by \cite{wong2020overfitting}, adversarial training is susceptible to overfitting. In our ResNet-18 experiments, we found that training for more than 100 epochs tended to do more harm than good, so in our experiments we train for either 80 or 100 epochs. For best results in this regime, we swept learning rates across the different optimizers in \texttt{pytorch-optimizer} \footnote{\url{https://github.com/jettify/pytorch-optimizer}.} so that baseline PGD training \cite{madry2018towards} (Algorithm \ref{algo:PGD}) provided good results at the end of the last epoch. We found that Yogi \cite{NIPS2018_8186} with a learning rate of 0.003 (batch size 128) worked well. We use weight decay 5e-4. We reduce the learning rate at two steps by factor 0.1x at 70\% and 90\% of the training epochs. The mixup-based models tended to be more robust to overfitting, so we train them for 100 epochs; the only model trained for 80 epochs is baseline PGD \cite{madry2018towards}. The hyperparameters were selected for the baseline PGD adversarial training: due to this, our implementation of the PGD baseline (using PreActResNet18) in table \ref{tablecifar10} is stronger than the WideResNet reported by \cite{madry2018towards}.

In a second implementation, for experiments using WideResNet-34 (which is a far larger network computationally) in table \ref{tablecif10wide}, we use the project provided by \cite{wong2020overfitting} with similar hyperparameters (10-step PGD of step size 2/255, 200 epochs with dropped learning rate at 100 epochs, weight decay 5e-4, SGD optimizer with momentum 0.9, learning rate 0.1).

When reading the data, we use standard baseline augmentations (same as \cite{madry2018towards}): randomly flip left/right, and translate +/- 4 pixels, and normalize inputs by the 3-channel mean and std. dev. of the training set. \cite{wong2020fast} argues that ``1-step PGD'' (FGSM with uniform initialization $U(-\varepsilon,\varepsilon)$) works reasonably well for robustness, but we found that more steps provide additional benefits in robust accuracy and stability, so we train RN-18 against 5-step PGD with step size 4/255, and WRN-34 against 10-step PGD with step size 2/255. 

For all of our evaluations of Mixup, we initialize $\lambda$ from the symmetric Beta distribution $B(\alpha,\alpha)$ parameterized by $\alpha = 0.5$. In our experiments with learned $\lambda$, as parametrized in Algorithm \ref{algo:ouroptimizeratio}, we find that the initialization of $\lambda_{init}$ is somewhat important. It seems more valuable to initialize from interpolation positions nearer to original training datapoints using $B(\alpha,\alpha)$ rather than a uniform initialization $\lambda \sim U(0,1)$. For this reason we can also impose a clipping $\kappa > 0.5$, which would mean there are regions of the space perfectly between two datapoints that may never be learned from, but would help bias the optimization towards larger values, which are expected to be nearer to the real datapoints. We use $\kappa = 0.65$ in our experiments; the optimization is not very sensitive to the exact value of $\kappa$ and it seems less important than the initialization $\lambda_{init}$. To take steps we use projected gradients (sign of the gradient, multiplied by a step sized so that cumulatively a noticeable change in $\lambda$ could be reached).

As an aside, we note that our approach is compatible with the Manifold Mixup \cite{verma2019manifold} approach, mixing hidden features $x_i$ instead of input images. Unfortunately, we found its benefits in our robust optimization framework to be nearly negligible: it affects the resulting robust accuracy by no more than 0.1\%.

\subsection{Ablations \& Evaluation}

We start from the baseline PGD training \cite{madry2018towards} described in Algorithm \ref{algo:PGD}. Subsequently, we reproduce the two baseline approaches discussed in Algorithm \ref{algo:baseline1} \cite{wong2020overfitting} and Algorithm \ref{algo:interpadvtrain} \cite{lamb2019interpolated}, and then build our optimization framework. To measure the impact of optimizing the mixup ratio (algorithm \ref{algo:ouroptimizeratio}), in some experiments we freeze the value $\lambda$ after sampling from $B(\alpha,\alpha)$; otherwise in our full method we allow it to be optimized. We also measure the effect of the independently optimized left and right $\delta$ (versus Shared $\delta$).

For robust evaluation we follow the advice and guidelines of \cite{carlini2019on}. Note that we do not resort to any test-time tricks and at inference time our model is the same as any standard network. As a result, our primary concern would be obfuscated gradients \cite{athalye2018obfuscated}. We evaluate our attack against PGD adversaries with 20 iterations (PGD20) and with 100 iterations (PGD100), both using step size 2/255. The PGD-20 attacker is the same as \cite{madry2018towards} (step size 2/255, no restarts). We also make use of the recent AutoAttack toolbox \cite{croce2020reliable} which includes the black-box score-based (sample-querying, gradient-free) adversary Square \cite{andriushchenko2019square} which would be able to overcome obfuscated gradients, if they were a problem.

\subsection{Ablation of Algorithms \ref{algo:ours} and \ref{algo:ouroptimizeratio}}

We report results on CIFAR-10 in table \ref{tablecifar10} and results on CIFAR-100 in table \ref{tablecifar100}. All numbers are accuracies (percent correct out of 100, higher is better). AA refers to the AutoAttack toolbox from \cite{croce2020reliable}, which for table \ref{tablecifar10} we run in cheap mode, which uses fewer steps and samples. Despite being called cheap, we find its effectiveness to be close to the full attack -- for our CIFAR10 model in the last row of table \ref{tablecifar10}, we list the cheap score as 48.7\%; in our evaluation we found the full score to be 48.6\% which is very comparable: the full evaluation on our model did not reveal anything new. In most cases the black-box Square attack \cite{andriushchenko2019square}, the fourth attack in the AutoAttack ensemble and the only gradient-free attacker, found \textit{no} additional adversary images. Since the preceding 3 attacks were gradient-based, this points towards the implication that the gradients were not interfered with.
\begin{table}[]
\centering
\caption{Ablation Experiments, CIFAR-10 Accuracy, ResNet18}
\label{tablecifar10}
\begin{tabular}{|l|c|c|c|l|}
\hline
\diagbox[innerwidth=1.8cm]{Train}{Test}          & pristine & PGD20 & PGD100 & AAch \\ \hline
Attack                   & 83.3\%     & 50.2\%  & 49.8\%   & 47.3\%       \\ \hline
Mix-then-Attack          & 78.5\%  & 52.1\%   & 51.9\%    &  47.1\%        \\ \hline
Attack-then-Mix          & 82.5\%     & 49.9\%  & 49.7\%   &  46.6\%       \\ \hline
Ours, Fr. $\lambda$, Sh. $\delta$ & 82.3\%     & 51.5\%     & 51.1\%       &  47.6\%     \\ \hline
Ours, Frozen $\lambda$           & 80.5\%     & 52.4\%  & 52.2\%   & 48.2\%     \\ \hline
Ours, Shared $\delta$           & 81.6\%     & 52.9\%  & 52.7\%   & 48.3\%         \\ \hline
Ours                     & 79.8\%     & 54.1\%  & 54.0\%    &  48.7\%       \\ \hline
\end{tabular}
\end{table}

\begin{table}[]
\centering
\caption{Ablation Experiments, CIFAR-100 Accuracy, ResNet18}
\label{tablecifar100}
\begin{tabular}{|l|c|c|c|l|}
\hline
\diagbox[innerwidth=3cm]{Train}{Test}          & pristine & PGD20 & PGD100 \\ \hline
Attack                   & 56.7\%     & 25.1\%  & 24.9\%             \\ \hline
Mix-then-Attack          & 50.3\%     & 28.6\%  & 28.5\%              \\ \hline
Attack-then-Mix          & 55.7\%     & 27.3\%  & 27.2\%             \\ \hline
Ours, Frozen $\lambda$, Shared $\delta$ & 55.1\% & 28.0\%    & 27.9\%              \\ \hline
Ours, Frozen $\lambda$           & 53.8\%    & 29.3\%  & 29.2\%         \\ \hline
Ours, Shared $\delta$           & 53.7\%     & 28.8\%  & 28.6\%             \\ \hline
Ours                     & 52.0\%     & 29.2\%  & 29.1\%            \\ \hline
\end{tabular}
\end{table}

\subsection{Geometrical Label Mixing}

We report results on CIFAR-10 in table \ref{tablecif10wide}, this time using the much larger WideResNet-34 network \cite{zagoruyko2016wide}. As a baseline, we start from the code and implementation of \cite{wong2020overfitting}. These models are evaluated against AutoAttack in ``standard'' mode with the source implementation as of September 2020. Again, there are no issues with obfuscated gradients, as the robust accuracy holds up well against the strong AutoAttack suite which includes the gradient-free black-box Square attack. This table shows that the geometrical mixed labeling did not provide a large benefit, which indicates that the concerns were unlikely during the adversarial optimization. In a high dimensional space, it is unlikely that initialized perturbations would coincide with the vector connecting two data points. This makes implementation of our approach easier, since the interpolation method of Algorithm \ref{algo:ours} is simpler.

\begin{table}[]
\centering
\caption{CIFAR-10 Accuracy, Geometric Label Mixing, WRN-34}
\label{tablecif10wide}
\begin{tabular}{|l|c|c|c|l|}
\hline
\diagbox[innerwidth=3cm]{Train}{Test}  &  pristine  &  PGD10  & AA \\ \hline




Baseline \cite{madry2018towards,wong2020overfitting}  &  86.20 \%  & 56.54 \%  &  51.99 \%  \\ \hline

Ours, Alg. \ref{algo:ours} + \ref{algo:ouroptimizeratio}  & 85.11 \%  &  58.31 \%  & 52.64 \%       \\ \hline
Ours, Alg. \ref{algo:ours}+\ref{algo:betterlabelassignment}+\ref{algo:ouroptimizeratio}  & 85.04 \%  &  58.34 \%  & 52.69 \%       \\ \hline

\end{tabular}
\end{table}

\vspace{-1mm}
\section{Discussion}
\label{sec:discussion}
Our method provides significant improvements above the evaluated baselines, and holds up against the strong PGD100 and AutoAttack adversaries. There is a significant drop in robust accuracy from PGD100 to AutoAttack, but the general rank ordering of the approaches remains similar. The optimization criteria aims for a smooth robustness, so it is reasonable that the result is no more vulnerable to attack than the baseline PGD. Both CIFAR-10 and CIFAR-100 see about 4\% absolute improvement against PGD-100 over PGD adversarial training, which we refer to in our results as ``Attack'' without any mixup. It is important to note that this is a more significant \textit{relative} improvement for CIFAR-100 because the baseline accuracies are much lower. CIFAR-100 is a significantly more challenging dataset than CIFAR-10, both because there are more classes to confuse the classifier, and because there are fewer images-per-class (500 as opposed to 5000 in CIFAR-10). Because of the relatively smaller dataset of CIFAR-100 (fewer images-per-class), data augmentation becomes more important, so these Mixup strategies demonstrate their value more prominently. However the ablation effects are a little more muddled on CIFAR-100. In fact, our whole (non-ablated) model performed no better than the ablated model with frozen mixing ratio $\lambda$. This may be dataset dependent, depending on the distribution of pairwise distances between classes which is especially important under adversarial attacks. However, another probable reason behind this might be the specific formulation of the mixing optimization we tested (a shifted sigmoid initialized from a Beta distribution, but with no prior regularization during the optimization). Future work may focus on the formulation of the mixing distribution, but the results as in table \ref{tablecifar100} still demonstrate that our optimization method is stronger (by 0.6\% absolute improvement under 100-step PGD attack) than any of the other mixup-based baselines.

On CIFAR-10 in table \ref{tablecifar10} we see the strongest improvement of 2.1\% in PGD-100 and 1.4\% in AutoAttack over the strongest baseline. On both datasets, the ``Mix-then-attack'' is the strongest baseline. It is closely related to our method in the the adversarial optimization is performed in between datapoints $x_i$ and $x_j$, however our results show that it is valuable to be able to fully probe the space between datapoints with an integrated optimizer that can move the starting points ($x_i'$ and $x_j'$); otherwise the adversarial optimizer's clipping will prevent it from reaching potentially worse locations -- more incorrect locations that are desireable to learn from for maximal robustness.



\vspace{-2mm}
\section{Conclusion \& Future Work}
\label{sec:conc}
We have found improvements to adversarial training that addresses over-fitting in adversarial training that provides solid improvements in robust accuracy-under-attack over the baselines. The results are resilient to strong adversarial attacks, including black box gradient free attacks, demonstrating that we avoid gradient obfuscation. Our work demonstrates that when using data augmentation to improve the generalization of adversarial robustness, thoroughly adversarially backpropagating through the entire data augmentation formulation is important, because adversarial trained networks need to learn from the worst possible sample points.

We focused on aiming for maximum robustness by training only against the worst-case adversarial loss in each formulation. In future work we could attempt to explore the tradeoff between robust accuracy and natural (pristine) accuracy using either an interpolated adversarial + pristine combination of Interpolated Adversarial Training \cite{lamb2019interpolated} or more sophisticated balanced reformulations along the line of TRADES \cite{zhang2019theoretically}. Our formulation of the constrained optimization for the mixing ratio $\lambda$ used a sigmoid for clipping between 0.5 and 1, but future work could use or learn a better prior for $\lambda$, perhaps using a formulation like AdaMixUp \cite{guo2019mixup}. Any architecture can be trained with our approach, and as the authors of \cite{madry2018towards} found, larger networks with higher learning capacity tend to be able to achieve higher robust accuracies. Our approach could be combined with larger robustness-oriented architectures like RobNets \cite{guo2019meets} for further gains in robust accuracy.




\section*{Acknowledgment}
This material is based upon work supported by the Defense Advanced Research Projects Agency (DARPA). The views, opinions and/or findings expressed are those of the author and should not be interpreted as representing the official views or policies of the Department of Defense or the U.S. Government.

{\small
\bibliographystyle{ieee_fullname}
\bibliography{mycitations.bib}

\begin{thebibliography}{10}\itemsep=-1pt

\bibitem{andriushchenko2019square}
Maksym Andriushchenko, Francesco Croce, Nicolas Flammarion, and Matthias Hein.
\newblock Square attack: a query-efficient black-box adversarial attack via
  random search.
\newblock 2020.

\bibitem{archambault2019mixup}
Guillaume~P. {Archambault}, Yongyi {Mao}, Hongyu {Guo}, and Richong {Zhang}.
\newblock Mixup as directional adversarial training.
\newblock {\em arXiv preprint arXiv:1906.06875}, 2019.

\bibitem{athalye2018obfuscated}
Anish {Athalye}, Nicholas {Carlini}, and David {Wagner}.
\newblock Obfuscated gradients give a false sense of security: Circumventing
  defenses to adversarial examples.
\newblock In {\em ICML 2018: Thirty-fifth International Conference on Machine
  Learning}, pages 274--283, 2018.

\bibitem{carlini2019on}
Nicholas {Carlini}, Anish {Athalye}, Nicolas {Papernot}, Wieland {Brendel},
  Jonas {Rauber}, Dimitris {Tsipras}, Ian~J. {Goodfellow}, Aleksander {Madry},
  and Alexey {Kurakin}.
\newblock On evaluating adversarial robustness.
\newblock {\em arXiv preprint arXiv:1902.06705}, 2019.

\bibitem{croce2020reliable}
Francesco {Croce} and Matthias {Hein}.
\newblock Reliable evaluation of adversarial robustness with an ensemble of
  diverse parameter-free attacks.
\newblock In {\em ICML 2020: 37th International Conference on Machine
  Learning}, 2020.

\bibitem{goodfellow2015explaining}
Ian~J. {Goodfellow}, Jonathon {Shlens}, and Christian {Szegedy}.
\newblock Explaining and harnessing adversarial examples.
\newblock In {\em ICLR 2015 : International Conference on Learning
  Representations 2015}, 2015.

\bibitem{guo2019mixup}
Hongyu {Guo}, Yongyi {Mao}, and Richong {Zhang}.
\newblock Mixup as locally linear out-of-manifold regularization.
\newblock {\em AAAI 2019 : Thirty-Third AAAI Conference on Artificial
  Intelligence}, 33(1):3714--3722, 2019.

\bibitem{guo2019meets}
Minghao Guo, Yuzhe Yang, Rui Xu, Ziwei Liu, and Dahua Lin.
\newblock When nas meets robustness: In search of robust architectures against
  adversarial attacks.
\newblock In {\em Proceedings of the IEEE/CVF Conference on Computer Vision and
  Pattern Recognition}, pages 631--640, 2020.

\bibitem{he2016identity}
Kaiming He, Xiangyu Zhang, Shaoqing Ren, and Jian Sun.
\newblock Identity mappings in deep residual networks.
\newblock In {\em European conference on computer vision}, pages 630--645.
  Springer, 2016.

\bibitem{hendrycks2019natural}
Dan {Hendrycks}, Kevin {Zhao}, Steven {Basart}, Jacob {Steinhardt}, and Dawn
  {Song}.
\newblock Natural adversarial examples.
\newblock {\em arXiv preprint arXiv:1907.07174}, 2019.

\bibitem{ilyas2019adversarial}
Andrew {Ilyas}, Shibani {Santurkar}, Dimitris {Tsipras}, Logan {Engstrom},
  Brandon {Tran}, and Aleksander {Madry}.
\newblock Adversarial examples are not bugs, they are features.
\newblock In {\em NeurIPS 2019 : Thirty-third Conference on Neural Information
  Processing Systems}, pages 125--136, 2019.

\bibitem{lamb2019interpolated}
Alex {Lamb} et~al.
\newblock Interpolated adversarial training: Achieving robust neural networks
  without sacrificing too much accuracy.
\newblock In {\em Proceedings of the 12th ACM Workshop on Artificial
  Intelligence and Security}, pages 95--103, 2019.

\bibitem{lee2020adversarial}
Saehyung {Lee}, Hyungyu {Lee}, and Sungroh {Yoon}.
\newblock Adversarial vertex mixup: Toward better adversarially robust
  generalization.
\newblock In {\em CVPR 2020: Computer Vision and Pattern Recognition}, pages
  272--281, 2020.

\bibitem{madry2018towards}
Aleksander {Madry}, Aleksandar {Makelov}, Ludwig {Schmidt}, Dimitris {Tsipras},
  and Adrian {Vladu}.
\newblock Towards deep learning models resistant to adversarial attacks.
\newblock In {\em ICLR 2018 : International Conference on Learning
  Representations 2018}, 2018.

\bibitem{mangla2020varmixup}
Puneet {Mangla}, Vedant {Singh}, Shreyas~Jayant {Havaldar}, and Vineeth~N
  {Balasubramanian}.
\newblock Varmixup: Exploiting the latent space for robust training and
  inference.
\newblock {\em arXiv preprint arXiv:2003.06566}, 2020.

\bibitem{pang2020mixup}
Tianyu {Pang}, Kun {Xu}, and Jun {Zhu}.
\newblock Mixup inference: Better exploiting mixup to defend adversarial
  attacks.
\newblock In {\em ICLR 2020 : Eighth International Conference on Learning
  Representations}, 2020.

\bibitem{NEURIPS2019_9015}
Adam Paszke et~al.
\newblock Pytorch: An imperative style, high-performance deep learning library.
\newblock In {\em Advances in Neural Information Processing Systems 32}, pages
  8024--8035. Curran Associates, Inc., 2019.

\bibitem{szegedy2014intriguing}
Christian {Szegedy}, Wojciech {Zaremba}, Ilya {Sutskever}, Joan {Bruna},
  Dumitru {Erhan}, Ian {Goodfellow}, and Rob {Fergus}.
\newblock Intriguing properties of neural networks.
\newblock In {\em ICLR 2014 : International Conference on Learning
  Representations (ICLR) 2014}, 2014.

\bibitem{tramer2020adaptive}
Florian Tramer, Nicholas Carlini, Wieland Brendel, and Aleksander Madry.
\newblock On adaptive attacks to adversarial example defenses.
\newblock {\em arXiv preprint arXiv:2002.08347}, 2020.

\bibitem{verma2019manifold}
Vikas {Verma}, Alex {Lamb}, Christopher {Beckham}, Amir {Najafi}, Ioannis
  {Mitliagkas}, David {Lopez-Paz}, and Yoshua {Bengio}.
\newblock Manifold mixup: Better representations by interpolating hidden
  states.
\newblock In {\em ICML 2019 : Thirty-sixth International Conference on Machine
  Learning}, pages 6438--6447, 2019.

\bibitem{wong2020fast}
Eric Wong, Leslie Rice, and J~Zico Kolter.
\newblock Fast is better than free: Revisiting adversarial training.
\newblock {\em arXiv preprint arXiv:2001.03994}, 2020.

\bibitem{wong2020overfitting}
Eric {Wong}, Leslie {Rice}, and Zico {Kolter}.
\newblock Overfitting in adversarially robust deep learning.
\newblock In {\em ICML 2020: 37th International Conference on Machine
  Learning}, 2020.

\bibitem{xie2020adversarial}
Cihang {Xie}, Mingxing {Tan}, Boqing {Gong}, Jiang {Wang}, Alan~L. {Yuille},
  and Quoc~V. {Le}.
\newblock Adversarial examples improve image recognition.
\newblock In {\em CVPR 2020: Computer Vision and Pattern Recognition}, pages
  819--828, 2020.

\bibitem{zagoruyko2016wide}
Sergey {Zagoruyko} and Nikos {Komodakis}.
\newblock Wide residual networks.
\newblock In {\em British Machine Vision Conference 2016}, 2016.

\bibitem{NIPS2018_8186}
Manzil Zaheer, Sashank Reddi, Devendra Sachan, Satyen Kale, and Sanjiv Kumar.
\newblock Adaptive methods for nonconvex optimization.
\newblock In S. Bengio, H. Wallach, H. Larochelle, K. Grauman, N. Cesa-Bianchi,
  and R. Garnett, editors, {\em Advances in Neural Information Processing
  Systems 31}, pages 9793--9803. Curran Associates, Inc., 2018.

\bibitem{zhang2018mixup}
Hongyi {Zhang}, Moustapha {Cisse}, Yann~N. {Dauphin}, and David {Lopez-Paz}.
\newblock mixup: Beyond empirical risk minimization.
\newblock In {\em ICLR 2018 : International Conference on Learning
  Representations 2018}, 2018.

\bibitem{zhang2019theoretically}
Hongyang {Zhang}, Yaodong {Yu}, Jiantao {Jiao}, Eric {Xing}, Laurent~El
  {Ghaoui}, and Michael {Jordan}.
\newblock Theoretically principled trade-off between robustness and accuracy.
\newblock In {\em ICML 2019 : Thirty-sixth International Conference on Machine
  Learning}, pages 7472--7482, 2019.

\end{thebibliography}
}

\end{document}